\documentclass[conference]{IEEEtran}
\IEEEoverridecommandlockouts
\usepackage{amsmath,amssymb,amsfonts}
\usepackage{algorithmic}
\usepackage{textcomp}
\usepackage{array}
\usepackage{multirow}
\usepackage{booktabs} 
\usepackage{graphicx}
\ifCLASSOPTIONcompsoc
    \usepackage[caption=false, font=normalsize, labelfont=sf, textfont=sf]{subfig}
\else
\usepackage[caption=false, font=footnotesize]{subfig}
\fi
\usepackage{threeparttable} 

\usepackage{amsthm}
\theoremstyle{definition}
\usepackage{graphicx}
\usepackage{textcomp}
\usepackage{xcolor}
\usepackage{algorithm}
\usepackage{multirow}
\usepackage{url}
\usepackage{hyperref}

\newtheoremstyle{upright}
  {3pt}   
  {3pt}   
  {\normalfont}  
  {}      
  {\bfseries} 
  {.}     
  { }     
  {}      

\theoremstyle{upright}
\newtheorem{definition}{Definition}

\DeclareMathOperator*{\argmin}{arg\,min}
\def\BibTeX{{\rm B\kern-.05em{\sc i\kern-.025em b}\kern-.08em
    T\kern-.1667em\lower.7ex\hbox{E}\kern-.125emX}}
    
\begin{document}


\title{Introducing Interval Neural Networks for Uncertainty-Aware System Identification \\ 
}

 \author{\IEEEauthorblockN{Mehmet Ali Ferah}
 \IEEEauthorblockA{\textit{Artificial Intelligence and Intelligent Lab.} \\
 \textit{Istanbul Technical University}\\
 Istanbul, Türkiye \\
 ferah23@itu.edu.tr}
 \and
 \IEEEauthorblockN{Tufan Kumbasar}
 \IEEEauthorblockA{\textit{Artificial Intelligence and Intelligent Lab.} \\
 \textit{Istanbul Technical University}\\
 Istanbul, Türkiye \\
 kumbasart@itu.edu.tr}\thanks{T. Kumbasar was supported by the BAGEP Award of the Science Academy.}}

\IEEEpubid{%
  \makebox[\columnwidth]{979-8-3315-1088-6/25/\$31.00 \copyright\
2025 IEEE\hfill}%
  \hspace{\columnsep}\makebox[\columnwidth]{}
}

\maketitle

\begin{abstract}
System Identification (SysID) is crucial for modeling and understanding dynamical systems using experimental data. While traditional SysID methods emphasize linear models, their inability to fully capture nonlinear dynamics has driven the adoption of Deep Learning (DL) as a more powerful alternative. However, the lack of uncertainty quantification (UQ) in DL-based models poses challenges for reliability and safety, highlighting the necessity of incorporating UQ. This paper introduces a systematic framework for constructing and learning Interval Neural Networks (INNs) to perform UQ in SysID tasks. INNs are derived by transforming the learnable parameters (LPs) of pre-trained neural networks into interval-valued LPs without relying on probabilistic assumptions. By employing interval arithmetic throughout the network, INNs can generate Prediction Intervals (PIs) that capture target coverage effectively. We extend Long Short-Term Memory (LSTM) and Neural Ordinary Differential Equations (Neural ODEs) into Interval LSTM (ILSTM) and Interval NODE (INODE) architectures, providing the mathematical foundations for their application in SysID. To train INNs, we propose a DL framework that integrates a UQ loss function and parameterization tricks to handle constraints arising from interval LPs.  We introduce novel concept "elasticity" for underlying uncertainty causes and validate ILSTM and INODE in SysID experiments, demonstrating their effectiveness.

\end{abstract}

\section{Introduction}
System identification (SysID) is the process of modeling dynamic systems based on data, where the system's future behavior depends on its historical dynamics \cite{ljung1999system}. Traditional SysID methods are well-developed but emphasize linear modeling \cite{dankers2016, zhou2006}, making them potentially inefficient for representing the nonlinear relationships observed in real-world systems. Integrating Deep learning (DL) into SysID has emerged as a powerful method, as DL can effectively capture complex mappings by modeling high-dimensional and nonlinear dynamics \cite{DLsytsid,kecceci2024novel}. Long Short-Term Memory (LSTM) \cite{LSTM} and Neural Ordinary Differential Equations (NODEs) \cite{node} networks are among the most widely used Neural Network (NN) architectures in SysID \cite{tuna2022deep,rahman2022neural}. Both have strengths that complement each other: LSTMs excel at discrete updates, while NODEs can also handle continuous dynamics effectively \cite{coelho2024enhancing}. 

The primary challenge with NNs is that their point predictions can be unreliable. To address this, Uncertainty Quantification (UQ) becomes crucial, as it allows for a more comprehensive evaluation of prediction reliability by quantifying the uncertainty associated with the model's outputs \cite{uq,uq2,guven2024exploring,koklu2024odyssey}. Instead of point predictions, UQ introduces Prediction Intervals (PIs), offering a range where the true outcome likely falls \cite{PI,PI2}. Popular methods for UQ typically model uncertainty in Learnable Parameters (LPs) by assuming they follow prior distributions \cite{gausslp}. Predefined distributions constrain model flexibility, potentially limiting generalization. Interval Neural Network (INN) provide an alternative approach for PIs that doesn’t rely on probabilistic assumptions \cite{baker1998universal}. Recent studies have explored the application of INNs across a wide range of fields \cite{intervalbound, intervalimprecise, SADEGHI2019338, oala2020interval}.

This paper presents a systematic method for constructing and learning INNs with the capability to perform UQ for SysID tasks. We propose constructing the INN by transforming the LPs of a pre-trained NN into interval LPs, without relying on probabilistic assumptions. Thus, by representing uncertainty via interval LPs and incorporating interval arithmetic operations throughout the layers, INNs can effectively capture the uncertainty. This approach enables the generation of Prediction Intervals (PIs), providing a measure of uncertainty in the predictions. We provide all the mathematical foundations on how to extend LSTM and NODE into Interval LSTM (ILSTM) and Interval NODE (INODE). To train the ILSTM and INODE that ensure PIs satisfy the desired coverage, we propose a DL framework that incorporates a UQ loss and parameterization tricks to eliminate the constraint arising from the interval LPs. The DL framework is designed to effectively transform LPs of a pre-trained NN into interval LPs while minimizing the width of the PIs, thereby balancing the trade-off between coverage accuracy and interval precision. To evaluate the UQ performances of ILSTM and INODE, we conduct a series of SysID experiments. The results show that both methods are effective in capturing uncertainty, with INODE demonstrating better performance in terms of coverage. To understand the underlying reasons, we introduce the concept of "elasticity," which provides a fresh perspective for analyzing uncertainty in LPs. The results of this study demonstrate that the proposed INNs effectively tackle SysID tasks while enabling UQ.

\section{Preliminaries on Interval Arithmetic}\label{interval_arithmetic}

\begin{definition}\label{def1}
For $x \in \mathbb{R}$, a \textit{crisp number} $a$ is a single, deterministic and finite value such that:
\[
x = a, \quad a \in \mathbb{R}
\]
\end{definition}

\begin{definition}\label{def2}
Let $\underline{a}, \overline{a} \in \mathbb{R}$ and $\underline{a} \leq \overline{a}$.  
An \textit{interval number} $\tilde{a} = [\underline{a}, \overline{a}]$ is defined as the closed subset of crisp numbers, defined in Definition \ref{def1}, given by \cite{interval_artihmetic}:
\[
\tilde{a} = [\underline{a}, \overline{a}] = \{ x \in \mathbb{R} \mid \underline{a} \leq x \leq \overline{a} \}.
\]
\noindent\textit{Special Case:}  
When $\underline{a} = \overline{a}$, the interval $[\underline{a}, \overline{a}]$ collapses to a single point $a$, representing a crisp number \cite{interval_artihmetic}.
\end{definition}

Given interval numbers \([\underline{a}, \overline{a}] \) and \( [\underline{b}, \overline{b}] \), the following operations are defined for interval arithmetic \cite{interval_artihmetic}:

\begin{itemize}
    \item Addition:  
$[\underline{a}, \overline{a}] + [\underline{b}, \overline{b}] = [\underline{a} + \underline{b}, \overline{a} + \overline{b}]$ 
    \item Subtraction:  
$[\underline{a}, \overline{a}] - [\underline{b}, \overline{b}] = [\underline{a} - \overline{b}, \overline{a} - \underline{b}]$ 
\item Multiplication:  
$[\underline{a}, \overline{a}] [\underline{b}, \overline{b}] = [\min(S), \max(S)],$ \label{intmul}  
\text{where } \noindent$S = \{\underline{a} \cdot \underline{b}, \underline{a} \cdot \overline{b}, \overline{a} \cdot \underline{b}, \overline{a} \cdot \overline{b}\}$.
\end{itemize}

We define an interval matrix with \(\underline{A}, \overline{A} \in \mathbb{R}^{m \times n}\), as follows:
\begin{align} \label{intMatrix}
[\underline{A}, \overline{A}] = \{ X \in \mathbb{R}^{m \times n} \mid \underline{A} \leq X \leq \overline{A} \}
\end{align}
Basic matrix operations, such as addition, subtraction, and multiplication, can be easily extended to interval matrices.
\begin{itemize}
    \item Addition and Subtraction:  Let \([\underline{A}, \overline{A}]\) and \([\underline{B}, \overline{B}]\) be as defined in \eqref{intMatrix}. The interval operations are defined as: 
\begin{equation} \label{matrixsum}
[\underline{A}, \overline{A}] + [\underline{B}, \overline{B}] = [\underline{A} + \underline{B}, \overline{A} + \overline{B}]
\end{equation}
\begin{equation} \label{matrixsubs}
[\underline{A}, \overline{A}] - [\underline{B}, \overline{B}] = [\underline{A} - \overline{B}, \overline{A} - \underline{B}]
\end{equation}
\item Multiplication: Let \([ \underline{A}, \overline{A} ]\) and \([ \underline{B}, \overline{B} ]\) as in \eqref{intMatrix} but with \(\underline{A}, \overline{A} \in \mathbb{R}^{m \times p}\) and \(\underline{B}, \overline{B} \in \mathbb{R}^{p \times n}\). It is defined as:
\begin{equation} \label{intmatrixMultiplication}
[ \underline{C}, \overline{C} ] = [ \underline{A}, \overline{A} ] [ \underline{B}, \overline{B} ]
\end{equation}
Here, the interval dot product is required. Let \([\underline{u}, \overline{u}]\) and \([\underline{v}, \overline{v}]\) be interval vectors with \(\underline{u}, \overline{u}, \underline{v}, \overline{v} \in \mathbb{R}^n\). 
\begin{equation} \label{intdot}
[\underline{u}, \overline{u}] \cdot [\underline{v}, \overline{v}] = [\min(S1, S2), \max(S1, S2)],
\end{equation}
where
\begin{equation}
\begin{split}
S1 &= \sum_{i=1}^{n} [\underline{u}(i) \overline{u}(i)] [\underline{v}(i), \overline{v}(i)] \\
S2 &= \sum_{i=1}^{n} [\underline{u}(i) \overline{u}(i)] [\underline{v}(i), \overline{v}(i)].
\end{split}
\end{equation}
\end{itemize}

\section{Learning NNs for SysID}\label{systemident}

Here, we briefly outline how the NNs are defined and learned for SysID problems. We assume having a dataset, \( \{(u(k), \hat{y}(k))\}_{k=1}^{K} \), with an input \( u(k) \) and an output $\hat{y}(k)$. To represent the dynamics, we aim to learn an NN as follows: 
\begin{equation}\label{systemid}
{y}(k) = g(x(k), \theta)
\end{equation}
with
\begin{equation}\label{NNinput}
x(k) = \begin{aligned}
& \big[ u(k-n_d), \, \dots, \, u(k-n_d-n_x), \\
& \quad y(k-1), \, \dots, \, y(k-n_y) \big]
\end{aligned}
\end{equation}
where \( n_x \) is input lag , \( n_y \) is output lag and  \( n_d \) is  dead time . 

To train the NN model within a DL framework, we first extract \( B = K - N \) trajectories with window length \( N \) to define the mini-batches via Algorithm-\ref{alg1}. Then, the model is trained with a DL optimizer in mini-batches of size \( mbs \). The optimization problem to minimized is defined as follows: 
\begin{equation}\label{loss}
\theta^* = \argmin_{\theta} \left\{ \frac{1}{BN} \sum_{m=1}^{B}\sum_{k=1}^{N} \left( \hat{Y}(m,k) - Y(m,k) \right)^2 \right\}
\end{equation}
During training and inference, the model operates in simulation mode, with \( x(k) \) updated after each prediction.

This paper uses LSTM and NODE networks to represent \(g(.)\) in \eqref{systemid}. All networks are implemented using feedforward NNs as layers. The remainder of this section provides a brief overview of handled feedforward NNs, LSTMs, and NODEs.

\begin{algorithm} [b]
\caption{Windowing Data}
\label{alg1}
\begin{algorithmic}[1]
\STATE \textbf{Input:}  \( \{(u(k), \hat{y}(k)\}_{k=1}^{K} \), \(N\)
\FOR{\( i = 1 \) \textbf{to} \( K - N \)}
   \STATE \( \{(U(k,:), \hat{Y}(k,:))\} = \{(u(k:k+N-1), \hat{y}(k:k+N-1))\} \)
\ENDFOR
\STATE \textbf{Output}:\(U,\hat{Y}\) 
\end{algorithmic}
\end{algorithm}
\subsection{Feedforward Neural Networks}\label{NNsec}
An \(n\)-layer NN is a function with an input \(x\) as follows:
\begin{equation}\label{NN}
y = g(x; \theta)
\end{equation}
with crisp LPs $\theta = \{\theta_i\}_{i=1}^{n}$. Here, \(g\) is defined as:
\begin{equation}\label{NNlayer}
g(x; \theta) = g_n(  g_{n-1}(g_1(x; \theta_1); \dots; \theta_{n-1}); \theta_n)
\end{equation}
The inference of $g_i$ is defined for \(x_i \in \mathbb{R}^{m}\) as follows:  
\begin{equation}\label{NNforward}
g_i(x_i; \theta_i) = \sigma_i(W_i x_i + b_i)
\end{equation}
with \(\sigma = \{ \sigma_i \}_{i=1}^{n}\) represents the activation functions and \(\theta_i = (W_i, b_i)\) consists \(W_i \in \mathbb{R}^{p \times m}\) and \(b_i \in \mathbb{R}^p\).

\subsection{Long Short Term Memory Network}\label{LSTM}
For a sampled input $x(k)$, an \(n\)-layer LSTM is defined : 
\begin{equation}\label{lstmstacked}
(h(k), c(k)) = g(x(k), h(k-1), c(k-1); \theta)
\end{equation}
where each layer is characterized by the input gate \(i(k)\), forget gate \(f(k)\), output gate \(o(k)\), hidden state \(h(k)\), and cell state \(c(k)\) at time step \(k\), which are defined as \cite{tuna2022deep}: 
\begin{equation}\label{lstm}
\begin{aligned} 
i_i(k) &= \sigma^{\text{sig}}(W_i^\textit{i} x_i(k) + U_i^\textit{i} h_i(k-1) + b_i^\textit{i}) \\
f_i(k) &= \sigma^{\text{sig}}(W_i^f x_i(k) + U_i^f h_i(k-1) + b_i^f) \\
o_i(k) &= \sigma^{\text{sig}}(W_i^o x_i(k) + U_i^o h_i(k-1) + b_i^o) \\
\tilde{c}_i(k) &= \sigma^{\text{tanh}}(W_i^c x_i(k) + U_i^c h_i(k-1) + b_i^c) \\
c_i(k) &= f_i(k) \odot c_i(k-1) + i_i(k) \odot \tilde{c}_i(k) \\
h_i(k) &= o_i(k) \odot \sigma^{\text{tanh}}(c_i(k))
\end{aligned}
\end{equation}
with 
$\theta_{i} = (W_{i}^{f}, U_{i}^{f}, b_{i}^{f}, W_{i}^{\textit{i}}, U_{i}^{\textit{i}}, b_{i}^{\textit{i}}, W_{i}^{c}, U_{i}^{c}, b_{i}^{c}, W_{i}^{o}, U_{i}^{o}, b_{i}^{o})$, where \(i\), \(f\), \(o\) and \(c\) indicate gates. In \eqref{lstm},  \(\sigma^{\text{sig}}\) \ and \(\sigma^{tanh}\) are the sigmoid and tanh activation functions, respectively. For point prediction, the final layer is defined as:
\begin{equation}\label{LSTM_N}
y(k) = W_n h(k) + b_n
\end{equation}

\subsection{Neural Ordinary Differential Equations Networks}\label{NODE}
Given the input \(({x}(t), t)\), state \(h(t)\) at time \(t\), the \(n\)-layer NODE is defined as \cite{node}:
\begin{equation}\label{nodecont}
\frac{d{h}(t)}{dt} = g({x}(t), t; \theta)
\end{equation}
To obtain \({h}(T)\) at time \(T\), we solve the following integral:
\begin{equation}\label{nodesolve}
{h}(T) = {h}(t_0) + \int_{t_0}^{T} g({x}(t), t; \theta) \, dt
\end{equation}

In this study, we apply Euler's method and rewrite \eqref{nodesolve} as \cite{zhang2019anodev2}:
\begin{equation}\label{nodediscrete}
{h}(k) = {h}(k-1) + g({x}(k); \theta)
\end{equation}
which is defined with $g_i$ \((i = 1, \dots, n-1)\) as:
\begin{equation}\label{nodelayer}
g_i(x_i(k); \theta_i) = \sigma_i(W_i x_i(k) + b_i)
\end{equation}
and the final layer \(i = n\) with :
\begin{equation}\label{nodelastlayer}
y(k)  = W_n x_n(k) + b_n
\end{equation}





\section{Interval NNs for UQ: Structure and Inference}
This section presents INNs for SysID, which can generate PIs. We first extend the feedforward NN into its INN counterpart as it is used to define layers of ILSTM and INODE in detail. Then, we define the equations of ILSTM and INODE for PI generation. We base our INN construction approach on a pre-trained NN with  LP \(\theta^*\). 

\subsection{Feedforward Interval Neural Network Construction}\label{INN}

A pre-trained NN is transformed into an INN by converting its LP \(\theta^*\) into an interval-valued set \(\tilde{\theta}\) as follows:
\begin{equation}\label{interval_theta}
\tilde{\theta}_i=[\underline{\theta}_i, \overline{\theta}_i ] = [\theta_i^* - \underline{\Delta}_i, \theta_i^* +\overline{\Delta}_i]
\end{equation}
where \(\tilde{\Delta} = \{\tilde{\Delta}_1, \dots, \tilde{\Delta}_n\}\) are the LPs of the INN, defined as \(\tilde{\Delta_i} = [\underline{\Delta_i}, \overline{\Delta_i}]\). Note that we must ensure \(\underline{\theta_i} < \overline{\theta_i}\) as required in Definition-\ref{def2}. 

Now, let us define the inference of the INN with \(\tilde{\theta}\).
\begin{equation}\label{intervalnn}
\tilde{y} = \tilde{g}(x;\tilde{\theta})
\end{equation}
Here, the INN uses the interval arithmetic operations presented in Section \ref{interval_arithmetic}. The interval version of \eqref{NNlayer} is:
\begin{equation}\label{intervallayer}
\tilde{g}(x; \tilde{\theta}) = g_n( g_{n-1}(  g_1(x; \tilde{\theta}_1); \dots);\tilde{\theta}_{n-1}); \tilde{\theta}_n)
\end{equation}
and more specifically, for layer \(i\), by defining \(\tilde{W}=[\underline{W},\overline{W}]\) and \(\tilde{b}=[\underline{b},\overline{b}]\), the forward pass is as follows:
\begin{equation}\label{intervalforward}
g_i(\tilde{x}_i; \tilde{\theta}_i) = \sigma_i( 
\tilde{W}_{i} \tilde{x}_i + \tilde{b}_{i})
\end{equation}

The output of each layer is an interval number, thus the last layer outputs an interval defined as $\tilde{y}=[\underline{y},\overline{y}]$. Note that, given  a crisp input to the first layer, one can use the special condition in Definition \ref{def2} and write \([\underline{x}_1,\overline{x}_1] = x\).

\subsection{Inference of Interval LSTM}\label{ILSTM}
This section introduces ILSTM based on the INN design. Thus, we extend \eqref{lstm} and \eqref{LSTM_N} by defining \(\tilde{\theta}\) as in \eqref{interval_theta}. We obtain:

\begin{equation}\label{ilstmi}
\begin{aligned}
\tilde{i}_i(k) &= \sigma^{sig}(\tilde{W}_i^\textit{i} \tilde{x}_i(k) + \tilde{U}_i^\textit{i} \tilde{h}_i(k-1) + \tilde{b}_i^\textit{i}) \\
\tilde{f}_i(k) &= \sigma^{sig}(\tilde{W}_i^f \tilde{x}_i(k) + \tilde{U}_i^f \tilde{h}_i(k-1) + \tilde{b}_i^f) \\
\tilde{o}_i(k) &= \sigma^{sig}(\tilde{W}_i^o \tilde{x}_i(k) + \tilde{U}_i^o \tilde{h}_i(k-1) + \tilde{b}_i^o) \\
\tilde{\tilde{c}}_i(k) &= \sigma^{tanh}(\tilde{W}_i^c \tilde{x}_i(k) + \tilde{U}_i^c \tilde{h}_i(k-1) + \tilde{b}_i^c) \\
\tilde{c}_i(k) &= \tilde{f}_i(k) \odot \tilde{c}_i(k-1) + \tilde{i}_i(k) \odot \tilde{\tilde{c}}_i(k) \\
\tilde{h}_i(k) &= \tilde{o}_i(k) \odot \sigma^{tanh}(\tilde{c}_i(k))
\end{aligned}
\end{equation}
Then, we can define the overall ILSTM network as:
\begin{equation}\label{ILSTM_N-1}
(\tilde{h}(k), \tilde{c}(k)) =  
 \tilde{g}(\tilde{x}(k), \tilde{h}(k-1),
 \tilde{c}(k-1); \tilde{\theta})
\end{equation}
with
\begin{equation}\label{ILSTM_N}
\tilde{y}(k) = \tilde{W}_n \tilde{h}(k) + \tilde{b}_n.
\end{equation}

As the aim is to generate a PI around the output of the pre-trained LSTM, we define the input \(\tilde{x}(k)\) in the INN by updating its output and states by using the pre-trained LSTM's output and state as follows:

\begin{equation}\label{centerh}
[\underline{h}(k-1), \overline{h}(k-1)] = [h(k-1), h(k-1)]
\end{equation}
\begin{equation}\label{centerc}
[\underline{c}(k-1), \overline{c}(k-1)] = [c(k-1), c(k-1)]
\end{equation}
\begin{equation}\label{centerylstm}
[\underline{y}(k-1), \overline{y}(k-1)] = [y(k-1), y(k-1)]
\end{equation}

\subsection{Inference of Interval NODE}\label{INODE}
We construct INODE similar to ILSTM by $\tilde{\theta}$ using \eqref{interval_theta}. The equations of INODE are expressed as follows:
\begin{equation}\label{intervalnode}
\tilde{h}(k+1) = \tilde{h}(k)+ \tilde{g}(\tilde{x}(k); \tilde{\theta})
\end{equation}
For layers \(i = 1, \dots, n-1\), we define:
\begin{equation}\label{intervalnodelayer}
\tilde{h}_i(k+1) = \tilde{h}_i(k) + \sigma_i(\tilde{W}_i\tilde{x}_i(k)  + \tilde{b}_i)
\end{equation}
while for the final layer \(i = n\), we define:
\begin{equation}\label{intervalnodelast}
\tilde{h}_n(k+1) = \tilde{h}_n(k) + (\tilde{W}_n \tilde{x}_n(k) + \tilde{b}_n)
\end{equation}

As we have done in ILSTM, to generate a PI around the output of the pre-trained NODE,  we update at each step INN output with:
 \begin{equation}\label{nodecenter}
[\underline{y}(k-1),\overline{y}(k-1)] = [y(k-1),y(k-1)]
\end{equation}

\section{Learning Interval NNs for UQ}

In this section, we present a DL framework to train the presented ILSTM and INODE networks for UQ. To train the INN, $L_{\text{RQR-W}}$ loss is used \cite{pouplin2024relaxed}. The problem to be minimized at every epoch is defined as: 
\begin{equation}\label{rqrw_loss} 
\tilde{\Delta}^* =\argmin_{\tilde{\Delta}} \left\{ \frac{1}{BN}  \sum_{m=1}^{B}\sum_{k=1}^{N} L_{\text{RQR}} +\lambda L_{\text{W}} \right\}
\end{equation}
where \(\alpha\) is the desired coverage to be captured by the PIs. Here, $L_{\text{RQR}}$ is the loss related to coverage and is defined as:
\begin{equation}\label{rqr_loss}
L_{\text{RQR}} = 
\begin{cases} 
\alpha \kappa & \text{if } \kappa \geq 0 \\
(\alpha - 1)\kappa & \text{if } \kappa < 0
\end{cases} \\
\end{equation}
where 
\begin{equation}
    \kappa = (\hat{Y}(m,k) - \underline{Y}(m,k))(\hat{Y}(m,k) - \overline{Y}(m,k)).
\end{equation}
The \(L_{\text{W}}\) term penalizes the PI width to enforce narrower bounds, weighted by the hyperparameter \(\lambda\). $L_{\text{W}}$ is defined as:
\begin{equation}\label{w_loss}
L_{\text{W}} = {(\overline{Y}(m,k) - \underline{Y}(m,k))^2}/{2}
\end{equation}

The training procedure for the INN is outlined in Algorithm-2
\footnote{MATLAB implementation. [Online]. \url{https://github.com/modifayd/INN\_UQ}} Here, we would like to underline that:  
\begin{itemize}
    \item During the training of the INN, we must ensure that all LPs \(\tilde{\Delta}\) result in \(\tilde{\theta}\) that satisfy the conditions of interval numbers (i.e. Definition-\ref{def2}). Thus, we must guarantee that \(\Delta_i \geq 0\). Yet, enforcing this constraint transforms \eqref{rqrw_loss} into a constraint optimization problem. Thus, as all the built-in DL optimizers are unconstrained ones, we perform the following parameterization tricks to eliminate the constraints:
\begin{equation}\label{trick}
\underline{\Delta}_i= \sigma_{\Delta}(\underline{\Delta}'_i), \quad
\overline{\Delta}_i= \sigma_{\Delta}(\overline{\Delta}'_i)
\end{equation}
where $\underline{\Delta}'_i$ and $\overline{\Delta}'_i$ are the new LPs, and \(\sigma_{\Delta}(.)\) is a function that ensures outputs remain strictly positive. In this paper, we propose two functions for $\sigma_{\Delta}$, which are absolute function $\sigma^{abs}$ and the ReLU function $\sigma^{ReLU}$.

\item  The initialization of \(\tilde{\Delta}\) is important as it might increase the training time. Given \(\theta^*\), the initialization of \(\tilde{\Delta}\) is performed for the hidden layers as follows: 
 \begin{equation}\label{rhidden}
        \{\tilde{\Delta}_1,...,\tilde{\Delta}_{n-1}\} = \{[{\theta_1^*}, {\theta_1^*}], \dots, [\theta_{n-1}^*, {\theta_{n-1}^*}]\} r_{\text{h}} 
 \end{equation}
where  $r_{\text{h}} \in [0,1]$ is the hidden layer uncertainty rate. The output layer is initialized with: 
\begin{equation}\label{rout}
\tilde{\Delta}_n\ = [{\theta_n^*},{\theta_n^*}] r_{\text{o}} 
  \end{equation}
where $r_{\text{o}} \in [0,1]$ is the 
is the output layer uncertainty rate. $r_{\text{h}}$ and $r_{\text{o}}$ are hyperparameters to be determined.
\end{itemize}

\begin{algorithm}[t]
\caption{Training INN}
\label{alg2}
\begin{algorithmic}[1]
\STATE \textbf{Input:} Training data: \( U,\hat{Y} \)
\STATE Mini-batch size: \(mbs\) 
\STATE Number of trajectory: \(B\) 
\STATE Number of epochs: \(E\) 

\STATE Coverage: $\alpha$, penalty: $\lambda$ 
\STATE Pre-trained NN: \( g \) with LP \( \theta^* \).
   \STATE \textbf{Output:} LP set \(  \tilde{\Delta}\) 

\STATE \textbf{Initialize \( \tilde{\Delta} \)}\hfill $\triangleright$ \textnormal{Eq}.\eqref{rhidden}-\eqref{rout}

\FOR {\(e=1\) to \(E\) }
    \FOR{each \(mbs\) in \(B\) }
        \STATE Select mini-batch {U(1:\(mbs\),1:N),Y(1:\(mbs\),1:N)}
         \STATE Construct X(1:\(mbs\),1:N) \hfill $\triangleright$ \(\textnormal{Eq}.\eqref{NNinput}\)    

        \STATE Compute Y$\leftarrow g($X(1:\(mbs\),1:N)$; \theta^*)$  \hfill $\triangleright$ Sec.  \ref{NN}
        \STATE Perform tricks and compute $\underline{\Delta}_i,\overline{\Delta}_i$ \hfill $\triangleright$ \(\textnormal{Eq}.\eqref{trick}\)

        \STATE Define \(\tilde{\theta}\) 
 \hfill $\triangleright$ \(\textnormal{Eq}.\eqref{interval_theta}\)

        \STATE Compute $\tilde{Y}$  $\leftarrow \tilde{g}($X(1:\(mbs\),1:N)$;\tilde{\theta})$ \hfill $\triangleright$ Sec. \ref{INN}     
        \STATE Compute loss: \( {L_{RQR-W}} \)\hfill $\triangleright$ \(\textnormal{Eq}.\eqref{rqrw_loss}\)    
        \STATE Compute gradients:
         \( {\partial {L}}/{\partial \tilde{\Delta}} \)     
      \STATE \text{Update $\tilde{\Delta}$ via Adam optimizer}
             
\ENDFOR
\ENDFOR
\STATE \(\tilde{\Delta}^* = \argmin({L_{RQR-W}}) \)
\RETURN $\tilde{\Delta}=\tilde{\Delta}^* $

\end{algorithmic}
\end{algorithm}

\section{Comparative Performance Analysis}

This section provides a comprehensive analysis of the learning performances of INNs on the SysID benchmark datasets: MR Damper \cite{wang2009identification}, Heat Exchanger \cite{DeMoorBLR1997}, and Hair Dryer \cite{ljung1999system}. All experiments were conducted in MATLAB\textsuperscript{\textregistered} and repeated with 5 different initial seeds for statistical analysis. The dataset configurations are set as given in Table \ref{tab:dataset_configs}.

\begin{table}[b]\label{table1}
\caption{Dataset Configurations and Hyperparameters}
\label{tab:dataset_configs}
\centering
\scriptsize 
\renewcommand{\arraystretch}{0.7} 
\setlength{\tabcolsep}{2pt} 
\begin{tabular}{l l|c|c|c c}
\toprule
\textbf{} & \textbf{} & \textbf{MR-Damper} & \textbf{Heat Exchanger} & \textbf{Hair Dryer} \\
\midrule
\multirow{4}{*}{\textbf{Config.}} 
& \# Samples & \( 3499 \)         & \( 4000 \)             & \( 1000 \)             \\
& Train-Val-Test & \( 54-13-33 \)        & \( 20-5-75 \)              & \( 40-10-50 \)              \\
& Normalization    & z-score               & min-max               & z-score               \\
& \( N \) & \( 40 \)           & \( 80 \)              & \( 30 \)              \\
\midrule
\multirow{4}{*}{\textbf{(I)NODE}} 
& \( n_x - n_d - n_y \) & \( 2 - 0 - 1 \)            & \( 0 - 0 - 3 \)           & \( 2 - 2 - 3 \)           \\
& \# Layers           & \( 2 \)                & \( 3 \)               & \( 3 \)               \\
& Hidden Layer Size   & \( 32 \)               & \( 35, 10 \)          & \( 40, 40 \)          \\
& \( r_{\text{out}} - r_{\text{hidden}} \) & \( 0.75-0.75 \)         & \( 1-1 \)             & \( 1-1 \)             \\
\midrule
\multirow{4}{*}{\textbf{(I)LSTM}} 
& \( n_x - n_d - n_y \) & \( 2 - 0 - 1 \)            & \( 2 - 0 - 1 \)           & \( 2 - 2 - 3 \)           \\
& \# Layers           & \( 3 \)                & \( 3 \)               & \( 3 \)               \\
& Hidden Layer Size    & \( 48, 48 \)           & \( 48, 48 \)          & \( 35, 35 \)          \\
& \( r_{\text{o}} - r_{\text{h}} \) & \( 1 - 0.2 \)            & \( 1 - 0.2 \)           & \( 1 - 0.2 \)           \\
\bottomrule
\end{tabular}
\end{table}

We present a dual-fold evaluation to analyze the proposed INNs.
\begin{itemize}
    \item Analyzing the coverage performances of proposed INNs for SysID tasks.
    \item Explaining how the proposed INNs quantify uncertainty within their LPs.
\end{itemize}

\subsection{Coverage Performance Analysis}

We first train the LSTM and NODE models as outlined in Section \ref{systemident}, with configurations summarized in Table \ref{tab:dataset_configs}. The mean testing performance, evaluated using the Root Mean Square Error (RMSE), is shown in Table \ref{table2}. The comparable performance across all datasets for each model enables an effective comparison within the INNs.

ILSTM and INODE models are constrcuted as described in Section \ref{ILSTM} and \ref{INODE}, respectively. We train the following INNs, each employing a different function within the parameterization tricks defined in \eqref{trick}:
\begin{itemize}
    \item ILSTM-1: ILSTM trained with  \(\sigma^{\text{ReLU}}\).
    \item ILSTM-2: ILSTM trained with  \(\sigma^{\text{abs}}\).
    \item INODE-1: ILSTM trained with  \(\sigma^{\text{ReLU}}\).
    \item INODE-2: ILSTM trained with  \(\sigma^{\text{abs}}\).
\end{itemize}
We trained all models using Algorithm-\ref{alg2} with configurations and hyperparameters provided in Table \ref{tab:dataset_configs}. The \(\Delta\) terms associated with reccurent weight \(U\) in both ILSTM-1 and ILSTM-2 were set to zero (i.e., no gradient flow) for MR Damper dataset.

We evaluated the UQ performances of the INNs by calculating the PI Coverage Probability (PICP) and PI Normalized Averaged Width (PINAW) metrics \cite{6581855}. We anticipate learning an INN that attains PICP close to the desired coverage with a low PINAW value. 

\begin{figure}[htbp!]
\centering
\resizebox{0.45\textwidth}{!}{\includegraphics{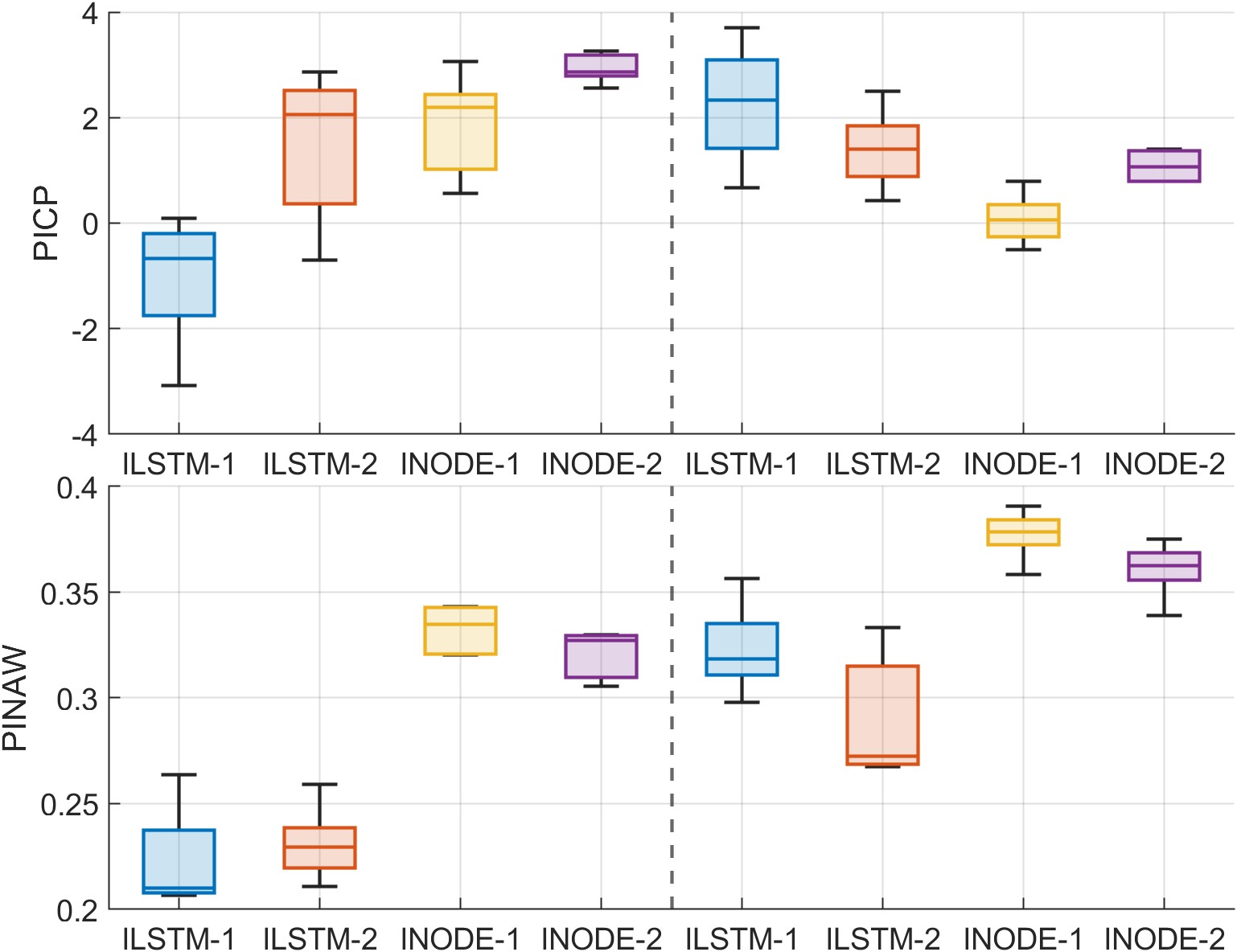}}
\caption{Heat Exchanger Dataset: UQ Performance Comparison}
\label{exchanger}
\end{figure}

\begin{figure}[htbp!]
\centering
\resizebox{0.45\textwidth}{!}{\includegraphics{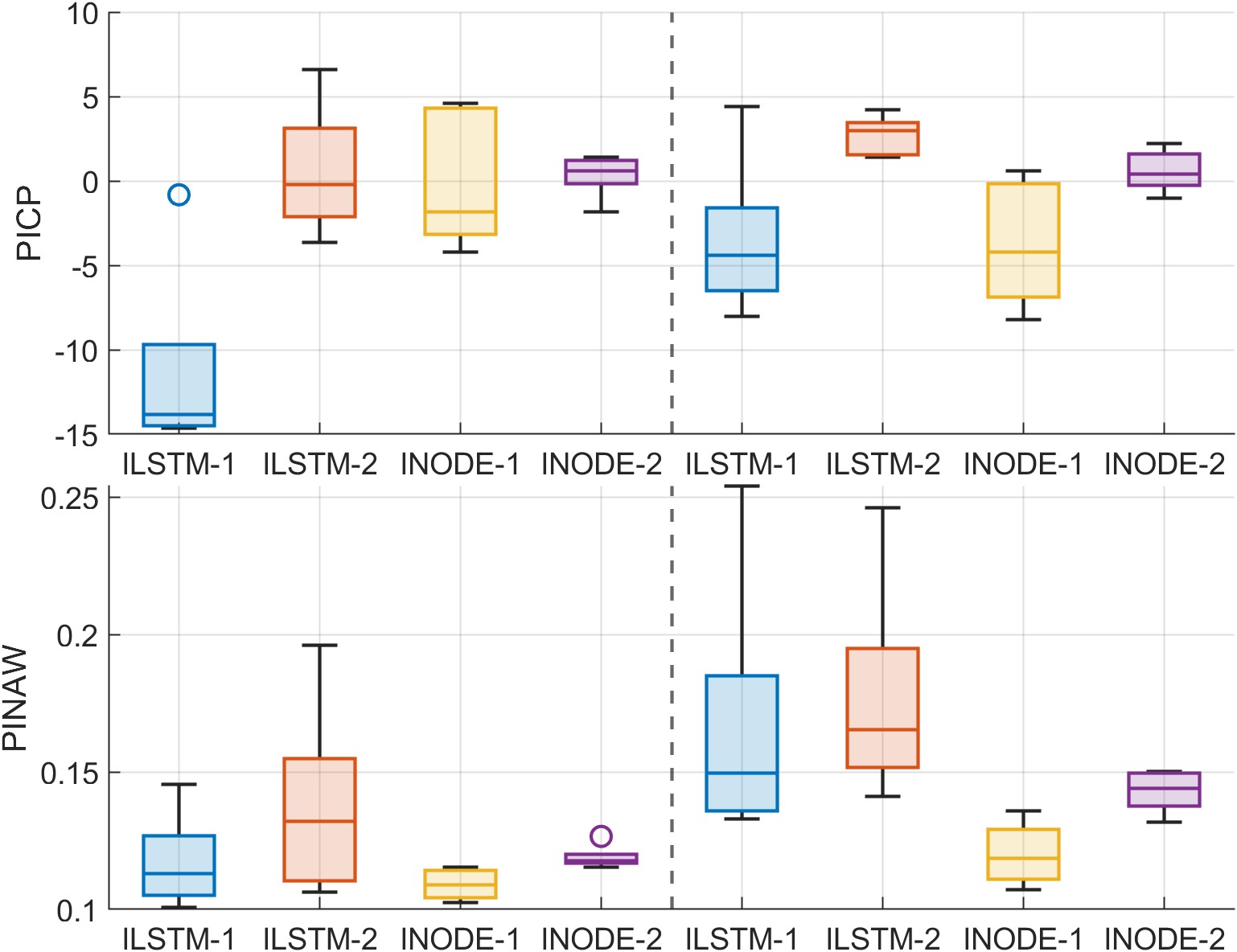}}
\caption{Hair Dryer Dataset: UQ Performance Comparison}
\label{dryer}
\end{figure}

\begin{figure}[htbp!]
\centering
\resizebox{0.45\textwidth}{!}{\includegraphics{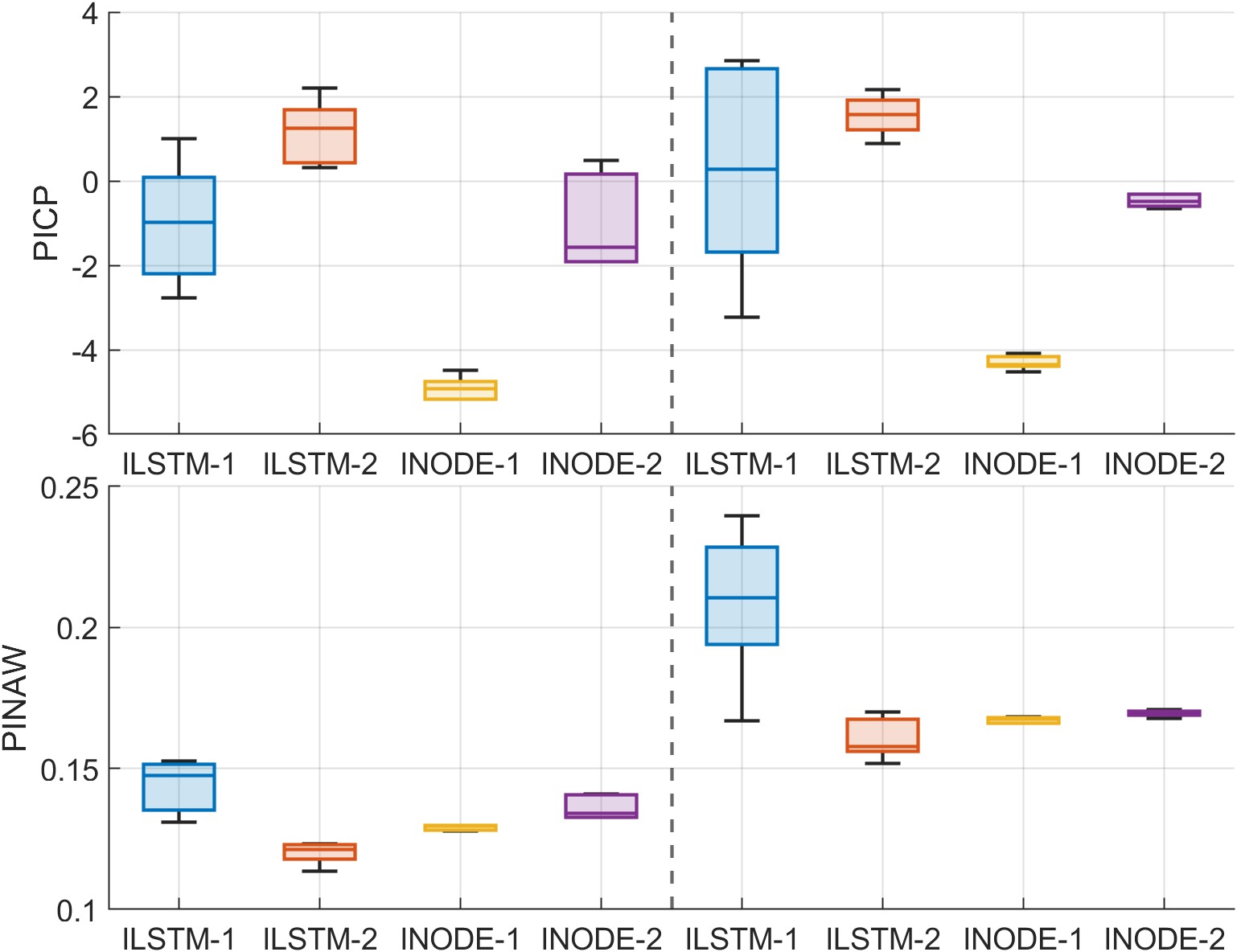}}
\caption{MR-Damper Dataset: UQ Performance Comparison}
\label{damper}
\end{figure}

Table \ref{tab:picp_pinaw} presents the PICP and PINAW values, expressed as mean $\pm$ standard deviation, for the desired coverage targets for these models are 90\% and 95\%. Fig. \ref{exchanger}-\ref{damper} present the box plots for each experiment. Here, the PICP values of box charts are generated by subtracting the calculated PICP from the target coverage. A dashed line in the middle indicates the coverage targets: 90\% coverage is represented before the dashed line, while 95\% coverage is shown after it. From the comparative results, we observe that:
\begin{itemize}  
    \item INODE-2 and ILSTM-2 show the best performance in terms of PICP. 
    \item The use of the absolute function instead of ReLU provides better target coverage.  
    \item When comparing ILSTM and INODE, the whiskers for PICP and PINAW are narrower for INODE, showing more consistent performance.  
\end{itemize}
In conclusion, INODE-2 shows the best UQ performance, balancing superior coverage with slightly wider intervals compared to ILSTM-2.

\begin{table}[t]
\caption{Testing RMSE Performance Comparision}
\label{table2}
\centering
\footnotesize 
\renewcommand{\arraystretch}{0.8} 
\setlength{\tabcolsep}{4pt} 
\begin{tabular}{l |c| c| c}
\toprule
\textbf{Dataset} & \textbf{LSTM} & \textbf{NODE} \\
\midrule
Heat Exchanger & 0.556\((\pm0.055)\) & 0.586\((\pm0.022)\) \\ 
MR-Damper & 5.883\((\pm0.192)\) & 6.659\((\pm0.044)\) \\ 
Hair Dryer & 0.100\((\pm0.008)\) & 0.093\((\pm0.006)\) \\ 
\bottomrule
\end{tabular}
\end{table}

\begin{table*}[t]
\caption{Testing Performance Metrics of ILSTM and INODE}
\label{tab:picp_pinaw}
\centering
\footnotesize 
\renewcommand{\arraystretch}{0.8} 
\setlength{\tabcolsep}{3pt} 
\begin{threeparttable} 
\begin{tabular}{l l|cc|cc|cc|cc}
\toprule
\textbf{Dataset} & \multirow{2}{*}{\textbf{Metric}} & \multicolumn{2}{c|}{\textbf{ILSTM-1}} & \multicolumn{2}{c|}{\textbf{ILSTM-2}} & \multicolumn{2}{c|}{\textbf{INODE-1}} & \multicolumn{2}{c}{\textbf{INODE-2}} \\
\cmidrule{3-4} \cmidrule{5-6} \cmidrule{7-8} \cmidrule{9-10}
& & \textbf{90\%} & \textbf{95\%} & \textbf{90\%} & \textbf{95\%} & \textbf{90\%} & \textbf{95\%} & \textbf{90\%} & \textbf{95\%} \\ 
\midrule
\multirow{2}{*}{Heat Exchanger} 
& PICP  & 88.95\((\pm1.11)\) & 97.25\((\pm1.04)\) & 91.47\((\pm1.30)\) & 96.40\((\pm0.68)\) & 91.84\((\pm0.88)\) & 95.08\((\pm0.43)\) & 92.94\((\pm0.25)\) & 96.09\((\pm0.26)\) \\ 
& PINAW & 22.34\((\pm2.16)\) & 32.33\((\pm1.92)\) & 23.07\((\pm1.59)\) & 29.02\((\pm2.64)\) & 33.23\((\pm1.00)\) & 37.74\((\pm1.06)\) & 32.72\((\pm1.02)\) & 36.25\((\pm1.20)\) \\ 
\midrule
\multirow{2}{*}{MR-Damper} 
& PICP  & 89.01\((\pm1.33)\) & 95.27\((\pm2.30)\) & 91.15\((\pm0.69)\) & 96.55\((\pm0.44)\) & 85.08\((\pm0.25)\) & 90.70\((\pm0.15)\) & 89.02\((\pm1.03)\) & 94.53\((\pm0.14)\) \\ 
& PINAW & 14.76\((\pm0.85)\) & 21.05\((\pm2.45)\) & 12.00\((\pm0.35)\) & 16.07\((\pm0.67)\) & 12.89\((\pm0.08)\) & 16.72\((\pm0.09)\) & 13.61\((\pm0.38)\) & 16.95\((\pm0.11)\) \\ 
\midrule
\multirow{2}{*}{Hair Dryer} 
& PICP  & 78.72\((\pm5.28)\) & 91.46\((\pm4.24)\) & 90.62\((\pm3.50)\) & 97.68\((\pm1.05)\) & 89.98\((\pm3.68)\) & 91.26\((\pm3.39)\) & 90.34\((\pm1.14)\) & 95.59\((\pm1.11)\) \\ 
& PINAW & 11.72\((\pm1.55)\) & 16.70\((\pm4.46)\) & 13.73\((\pm3.20)\) & 17.70\((\pm3.65)\) & 10.90\((\pm0.49)\) & 12.01\((\pm1.02)\) & 11.89\((\pm0.39)\) & 14.29\((\pm0.68)\) \\ 
\bottomrule
\end{tabular}
\begin{tablenotes}
\footnotesize
\item *PINAW values are scaled by 100.
\end{tablenotes}
\end{threeparttable}
\end{table*}

\begin{figure*}[t]   
    \centering
    \subfloat[INODE-1]{%
        \includegraphics[width=0.45\linewidth]{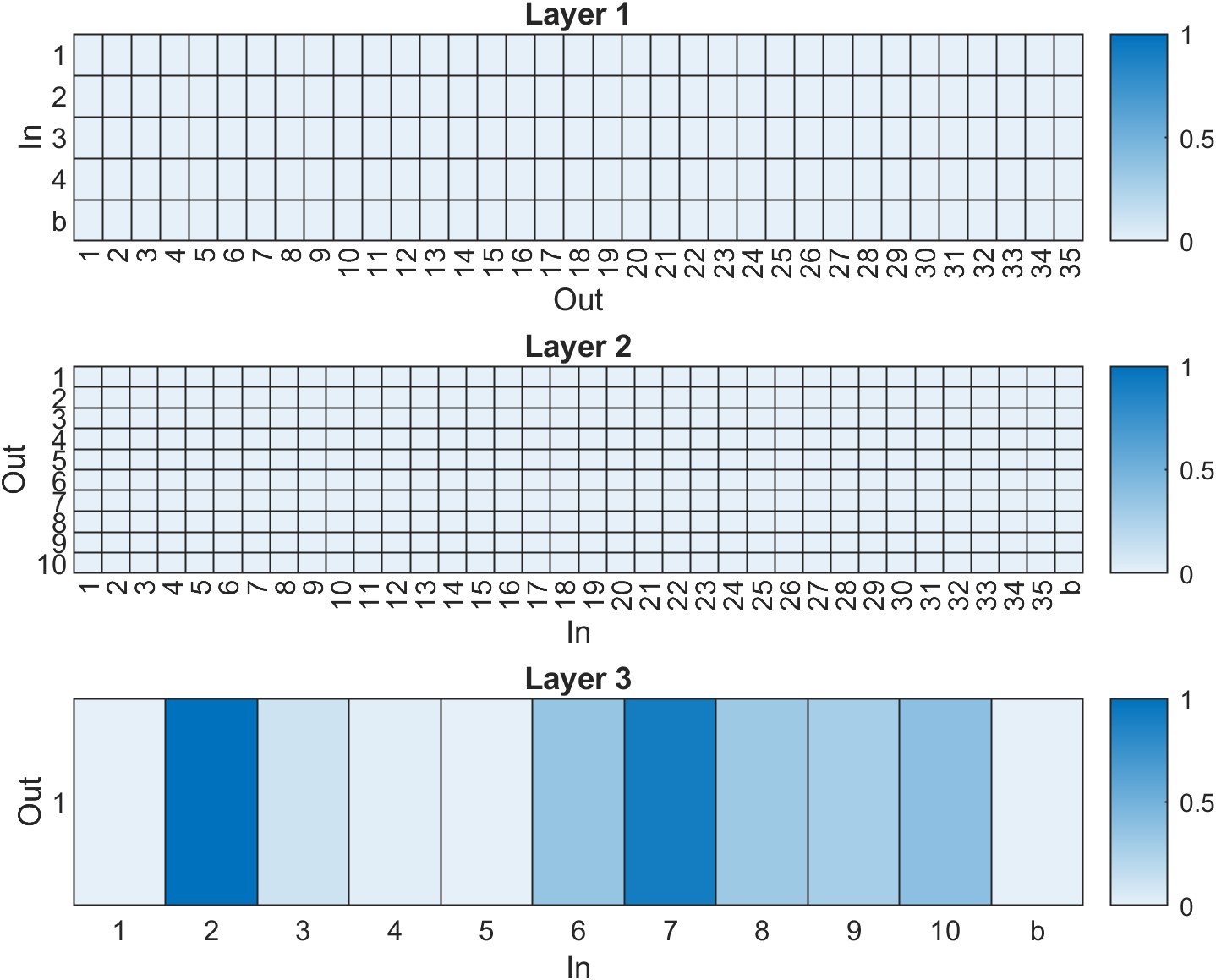}}
\hspace{0.6cm}
    \subfloat[INODE-2]{%
        \includegraphics[width=0.45\linewidth]{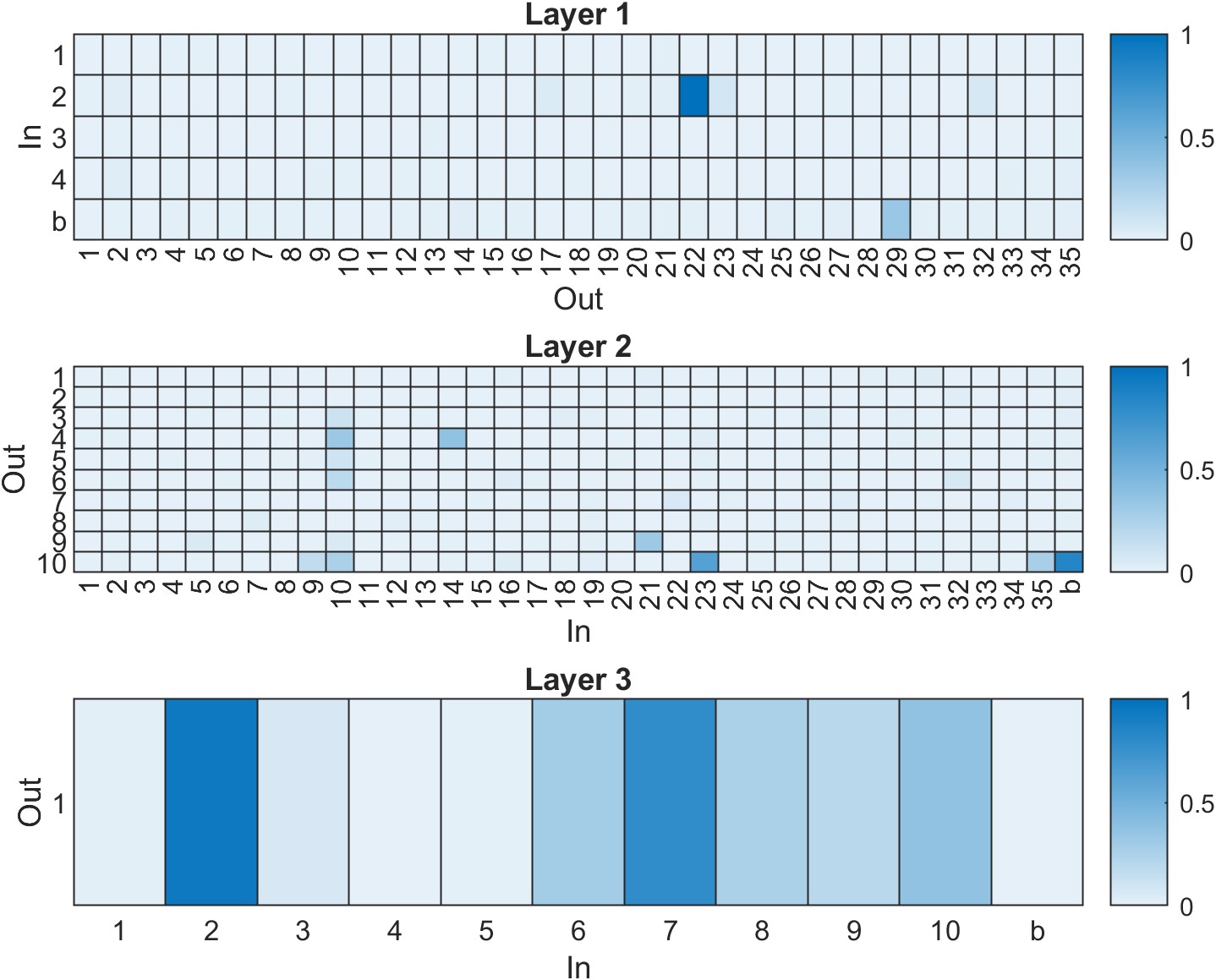}}
    \caption{Heatmaps of INN LPs from a SysID experiment on the Exchanger Dataset: (a) INDOE-1 and (b) INODE-2}   
     \label{heatmap}
\end{figure*}

\subsection{Analyzing the Uncertainty Representation}
This section analyzes how uncertainty is represented within the interval LPs \(\tilde{\theta}\) in the INN models, INODE-1 and INODE-2. The term "elasticity" is introduced to quantify the level of uncertainty in an INN parameter, defined as follows:
\begin{equation}\label{uncertainity}
    \bar{r}=\frac{\|\overline{\theta}-\underline{\theta}\|}{\|\theta^*\|}
\end{equation} 
 If $\bar{r}$ is close to 0, the uncertainty in the LP is minimal, indicating that the corresponding weight or bias is well-defined and does not contribute to uncertainty. On the other hand, if $\bar{r}$ is bigger than 1, the uncertainty is maximal, indicating that the parameter is highly uncertain. Fig. \ref{heatmap} illustrates the resulting $\bar{r}$ for each weight and bias, represented as a heatmap for the given model. We can observe that:  

\begin{itemize}
    \item For INODE-2, we can observe from Fig. \ref{heatmap}(b) that the elasticity of the weight corresponding to input 2 which is  $y(k-2)$ in layer 1 introduces uncertainty for output 22. In layer 2, the elasticity of the parameters corresponding to output 10 shows similar uncertainty. For the last layer, the elasticity is most pronounced in the parameters associated with outputs 2 and 7 of layer 2.    
    \item For INODE-1, the elasticity of the parameters in the last layer shows significant uncertainty as shown in Fig. \ref{heatmap}(a). The parameters associated with outputs 2 and 7 of layer 2 exhibit the highest elasticity, similar to the INODE-2.
\end{itemize}

Examining the elasticity of the parameters reveals which parts of the designed NN have a greater influence on the uncertainty of the output prediction. We observed that the parameters associated with the input \(y(k-2)\) exhibit the highest levels of uncertainty. It is also observed that INODE-1 introduces uncertainty mainly in the last layer.

\section{Conclusion and Future Work}

This paper presents a systematic approach for constructing INNs with the capability to perform UQ for SysID tasks. The widely used NODE and LSTM networks in SysID are extended to their interval counterparts, INODE and ILSTM. A DL framework is also developed for SysID tasks, integrating a UQ loss function and parameterization tricks to ensure robust and reliable performance. Experiments with four different configurations of INODE and ILSTM reveal that the proposed INNs successfully achieve target coverage with compact intervals. Among the configurations, INODE-2 demonstrates the best UQ performance, offering superior coverage while maintaining acceptable PI widths. Further analysis of the underlying dynamics of INODE-2, using \(\bar{r}\) values, have revealed that the \(\theta^*\) corresponding to the \(y(k-2)\) input resulted in the highest uncertainty. This observation highlights the model's sensitivity to the lagged output, offering valuable insights into the uncertainty contributions of specific parameters within the system.

In future research, we plan to extend the INN design to other NNs, such as CNNs, and explore their application to real-time tasks. Additionally, we aim to explore advancements in the interpretability of UQ for SysID tasks. 

\section*{Acknowledgment}

The authors acknowledge using ChatGPT to refine the grammar and enhance the English language expressions.

\bibliographystyle{IEEEtran}
\bibliography{cites}

\end{document}